\documentclass[lettersize,journal]{IEEEtran}
\IEEEoverridecommandlockouts
\usepackage{amsmath,amssymb,amsfonts}
\usepackage{textcomp}
\usepackage[hidelinks]{hyperref}
\usepackage{url}
\usepackage{cite}
\usepackage{placeins}
\usepackage{multirow}
\usepackage{stfloats}
\usepackage{booktabs}
\usepackage{balance}
\usepackage{graphicx}
\usepackage[caption=false,font=small,labelfont=rm,textfont=rm]{subfig}
\def\BibTeX{{\rm B\kern-.05em{\sc i\kern-.025em b}\kern-.08em
    T\kern-.1667em\lower.7ex\hbox{E}\kern-.125emX}}

\usepackage{fancyhdr}
\fancypagestyle{firstpage}{%
  \fancyhead{}
  
  \fancyfoot[C]{Submitted the IEEE Transactions on Cognitive and Developmental Systems, Copyright IEEE}
}

\begin{document}

\title{A Reinforcement Learning-Based Model for Mapping and Goal-Directed Navigation Using Multiscale Place Fields}


\author{Bekarys Dukenbaev\IEEEmembership{}, 
Andrew Gerstenslager\IEEEmembership{}, 
Alexander Johnson\IEEEmembership{},
Ali A. Minai \IEEEmembership{}
\thanks{B. Dukenbaev, A. Gerstenslager, and A. Johnson are with the Department of Computer Science,  University of Cincinnati, Cincinnati,
USA.}
\thanks{A.A. Minai is with the Department of Electrical and Computer Engineering,  University of Cincinnati, Cincinnati,
USA.}
}

\maketitle

\thispagestyle{firstpage}

\begin{abstract}
Autonomous navigation in complex and partially observable environments remains a central challenge in robotics. Several bio-inspired models of mapping and navigation based on place cells in the mammalian hippocampus have been proposed. This paper introduces a new robust model that employs parallel layers of place fields at multiple spatial scales, a replay-based reward mechanism, and dynamic scale fusion. Simulations show that the model improves path efficiency and accelerates learning compared to single-scale baselines, highlighting the value of multiscale spatial representations for adaptive robot navigation.
\end{abstract}
\IEEEpubidadjcol


\section{Introduction}
A unifying objective in autonomous robot navigation is to enable agents to learn from experience and reach specific goals with the adaptability and flexibility observed in animals. The challenge lies in building systems that can efficiently explore, map, and plan in complex environments without extensive prior knowledge. While traditional simultaneous localization and mapping (SLAM) methods rely on algorithmic reconstructions of geometric structure, they often struggle in unstructured environments with partial observability, sparse sensing, and scale, and incur substantial memory and compute costs \cite{thrun2005}. Mammals, in contrast, learn to navigate rapidly and adaptively in complex environments \cite{rosenberg2021labyrinth}, and form internal spatial representations and reach goals with limited experience, guided by the circuitry of the hippocampal complex using \textit{place cells} with location-specific activity \cite{okeefe1971hippocampus}, grid cells with spatially periodic activity \cite{Moser2008grid}, and replay/preplay mechanisms supporting swift value assignment and planning \cite{foster2006reverse,Dragoi2011,olafsdottir2016coordinated,olafsdottir2018hippocampal}. 

An important feature of the hippocampal system is the variation in the spatial scale of place fields along the dorsoventral axis \cite{Kjelstrup2008FiniteSO}. Computational models have typically adopted a single spatial resolution, leaving the functional implications of this multiscale organization underexplored. This paper introduces a new model using reinforcement learning that operationalizes multiscale representations, addressing key limitations of existing approaches by enabling dynamic scale integration. In the model, parallel layers of place fields are instantiated at distinct spatial resolutions and coupled to a lightweight value learner. At runtime, the system integrates scale-specific predictions through an adaptive fusion mechanism that selects whichever spatial resolution provides the clearest and most reliable directional reward structure at each decision point. The proposed architecture introduces two key innovations: (i) parallel multiscale neural populations and (ii) a dynamic scale weighting mechanism that uses differences in value-map structure across scales to stabilize policy updates without presuming a fixed role for any particular scale. Simulations across environments of varying size and complexity indicate that the multiscale model outperforms single-scale baselines in path efficiency, learning speed, and overall goal-reaching performance. Taken together, the results indicate that multiscale spatial representations improve RL-based navigation and provide a computational account of how hippocampal scale diversity can support adaptive behavior.


\section{Background and Motivation}
\subsection{The Neural Basis of Spatial Representation}
Understanding how animals represent and navigate space has been a central question in neuroscience and has provided key inspiration for bio-inspired robotics. The hippocampus and surrounding medial temporal lobe structures are critical to this capability, as evidenced by lesion studies showing severe impairments in rodents with hippocampal damage when performing spatial memory and navigation tasks\cite{Hales2014}.  

The discovery by O’Keefe and Dostrovsky of \emph{place cells} demonstrated that hippocampal neurons fire selectively when an animal occupies specific locations, supporting the idea of an internal \emph{cognitive map} of the environment \cite{Tolman1948,Okeefe1978}. Subsequent work revealed additional spatially tuned populations: \emph{head-direction cells} encode allocentric orientation \cite{taube1990head}, and \emph{boundary vector cells} respond to the distance and angle of environmental boundaries such as walls or barriers \cite{o1996geometric,barry2006boundary}. Together, these neural systems integrate sensory and self-motion cues to generate structured, multimodal representations that support flexible spatial memory and navigation.

\subsection{Multiscale Spatial Representations}
A hallmark of the hippocampal formation is the systematic gradient of spatial scale along the dorsoventral axis. Dorsal hippocampal cells exhibit small, high-resolution place fields that enable precise localization, while ventral cells display larger, diffuse fields that capture global spatial context \cite{jung1994comparison,Kjelstrup2008FiniteSO}. This \textit{multiscale} organization supports hierarchical spatial coding, with fine-scale representations supporting local navigation and coarse-scale representations enabling efficient long-range planning.

\subsection{Sequential Encoding and Plasticity}
Place cells encode not only location but also the temporal order of experiences. Spike-timing-dependent plasticity (STDP) strengthens connections between sequentially active cells, linking trajectories into coherent state-space representations \cite{Bi1998,Stachenfeld2017}. This mechanism explains why place fields respect environmental boundaries--cells on opposite sides of barriers are not experienced sequentially and thus remain weakly connected. STDP also contributes to reinforcement learning when coupled with neuromodulatory signals, allowing temporally ordered state sequences to be retrospectively associated with reward \cite{foster2006reverse,epsztein2022mental}.

\subsection{Replay and Preplay}
The hippocampus exhibits offline reactivation of spatial sequences during sharp-wave ripples, in which previously experienced or potential future trajectories are briefly replayed. In these events, the network briefly replays patterns of place-cell activity that represent past or potential future trajectories. \emph{Replay} refers specifically to the reactivation of sequences corresponding to previously experienced paths. These sequences may occur in forward or reverse order; forward replay supports memory consolidation, whereas reverse replay propagates reward information backward along experienced routes, strengthening earlier states' association with successful outcomes \cite{foster2006reverse,Lee2002,olafsdottir2016coordinated,olafsdottir2018role}.

In contrast, \emph{preplay} denotes the activation of trajectory-like sequences that occur \emph{before} the animal moves. These structured, prospective activations suggest that the hippocampus simulates candidate future paths, supporting planning and decision-making \cite{Dragoi2011}. Preplay is closely linked to vicarious trial-and-error behavior, where rodents pause at decision points and transiently evaluate alternative routes \cite{papale2016interplay,redish2016vicarious}.

Together, hippocampal replay and preplay correspond to several mechanisms central to modern reinforcement learning, including Monte Carlo tree search (MCTS) \cite{silver2016go}, rehearsal or ``dreaming'' \cite{hafner2023dreamerv3}, and eligibility traces \cite{sutton2018reinforcement}.

\subsection{Computational and Robotic Models}
Computational models of the hippocampus have long examined how place and boundary cells support spatial memory and navigation, typically using single-scale representations driven by attractor dynamics, Hebbian learning, boundary-vector cell inputs, and grid cells \cite{Burgess1994,samsonovich1997,tsodyks1999attractor,doboli2000latent,barry2006boundary,moser2008place}. Based on these models, several successful methods have been developed for navigation in complex environments with obstacles \cite{erdem2012goal,edvardsen2020navigation,alabi2020one,alabi2023rapid}. All of these approaches use some combination of memory replay, reward propagation, and path planning to find efficient paths, though there is considerable variation in detail. However, they are all based on place cells at a single scale of resolution. Erdem and Hasselmo have proposed a hierarchical model of navigation using multiscale place fields \cite{erdem2014biologically}, which was subsequently implemented with the RatSLAM model \cite{milford2004ratslam} in a visually-driven physical robot \cite{erdem2015hierarchical}. While this model uses multiscale forward replay for navigation, it does so to search through explicit goal-directed \textit{linear} trajectories in each episode, exploiting the longer reach of larger place fields to find a linear heading to the goal. This works well in small open environments but does not generalize to large and/or complex environments with obstacles. A complementary approach using multiscale place fields has been introduced to address this issue, demonstrating that larger fields enable rapid coarse exploration while smaller fields improve trajectory refinement near obstacles and goals \cite{Llofriu2015,scleidorovich2020multiscalenavigation,Scleidorovich2022}. However, this approach uses regular grids of fixed place fields. Neither approach supports the adaptive use of multiscale information. The present model addresses all these limitations by coupling self-organized parallel multiscale place-field populations to a reward-learning network that builds multiscale reward maps and performs online scale selection and adaptive fusion during action selection.



\section{Model Architecture}
The model comprises four core layers--Head Direction (HD) cells, Boundary Vector Cells (BVC), Place Cells (PC), and a Reward Network--instantiated in \emph{parallel} across multiple spatial scales. Each scale contains a full BVC–PC–Reward stack with its own tuning parameters, producing spatial codes and value estimates that range from fine to coarse resolution. These components behave in a manner broadly analogous to their biological counterparts: HD cells provide a global orientation signal, BVCs encode boundary geometry, place cells form location-specific representations, and the reward layer assigns value to visited states. This model builds on a simpler, single-scale model developed previously by our research group \cite{alabi2020one,alabi2022dissertation,alabi2023rapid}. 

\begin{figure*}[t]
    \centering
    \includegraphics[width=\textwidth]{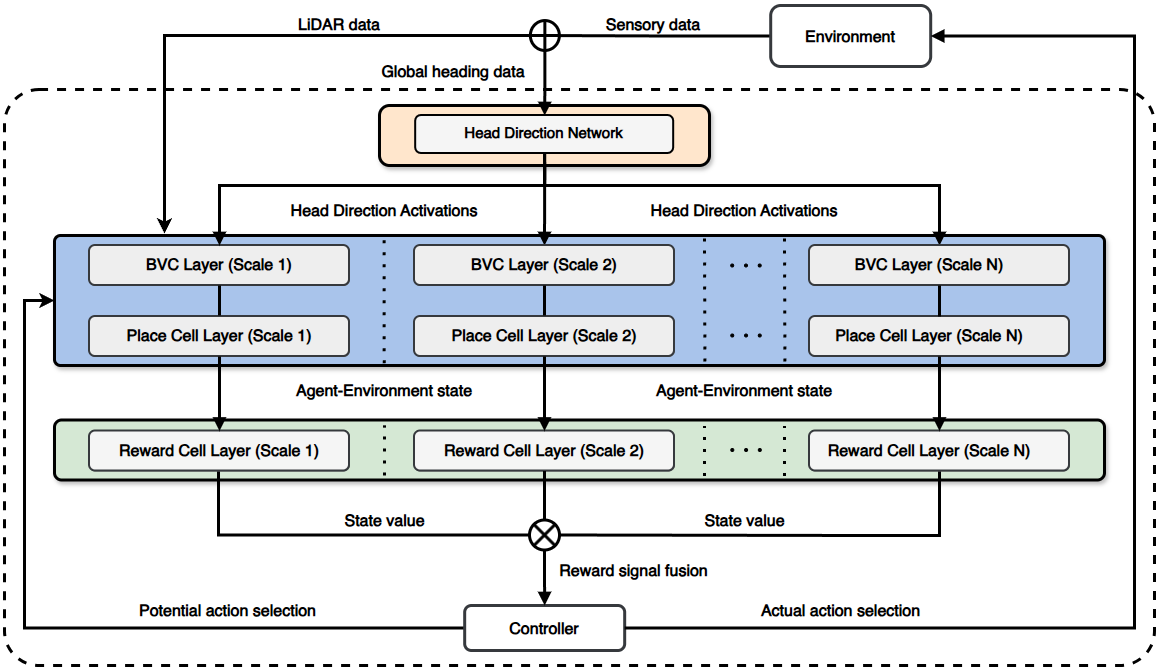}
    \caption{Multiscale system architecture integrating sensory inputs, neural processing layers, and decision-making components. The Head Direction Network processes global heading data, which modulates the activations of BVC and PC layers across multiple scales. The PC layer activations at each scale are then passed to their respective Reward Cell Layer, which returns a potential reward value for each possible heading. These reward values are then aggregated by a fusion module, after which a single, optimal action is chosen and executed.}
    \label{fig:img_multiscale_system_arch}
\end{figure*}

Unlike approaches that impose fixed place-field shapes (often Gaussian) or engineered layouts, all place fields here emerge dynamically through sensor-driven interaction with the environment. Multiscale structure is achieved by assigning each scale distinct BVC tuning widths \(\sigma_r\) (radial) and \(\sigma_\theta\) (angular), which determine place-field size and the spatial granularity of each scale’s reward map. Figure~\ref{fig:img_multiscale_system_arch} summarizes sensory inputs, neural processing at each scale, and downstream decision-making.

\subsection{Model Layers}
\subsubsection{Head Direction Cell Layer}
The HD layer encodes the agent’s allocentric heading. Each head-direction cell has a preferred direction in allocentric coordinates, and fires maximally when the agent’s heading matches its preferred direction, with a symmetric decrease in firing rate as the heading deviates from it. Thus, the activity of this layer represents a population coding of the agent’s directional heading relative to a fixed external reference, allowing it to maintain a global sense of orientation in the environment.

Formally, the firing rate of head-direction cell~$i$ is given by
\begin{equation}
\label{eqn_hdn}
    v_i^h \;=\; \mathbf{x}'\cdot 
    \begin{bmatrix}
        \cos\bigl(\theta_i^h + \theta_0\bigr) \\[4pt]
        \sin\bigl(\theta_i^h + \theta_0\bigr)
    \end{bmatrix},
\end{equation}
where $\theta_0$ denotes the heading angle of an \emph{anchor cue} (a fixed external reference), $\theta_i^h$ is the preferred allocentric direction of the $i$-th head-direction cell, and $\mathbf{x}' = [x'_0, x'_1]$ represents the agent’s instantaneous velocity in Cartesian coordinates \cite{erdem2014biologically}. The model uses $n_{\mathrm{hd}} = 8$ head-direction cells with preferred directions
\[
    \{\theta_d\}_{d=0}^{n_{\mathrm{hd}}-1}
    = \{0^\circ, 45^\circ, \dots, 315^\circ\},
\]
which we refer to as the set $\mathbf{\Theta}$ of canonical \emph{basis headings}.

\vspace{1.0em}

\subsubsection{Boundary Vector Cell Layer}
The BVC layer adapts the model of Barry \emph{et~al.}~\cite{barry2006boundary}, where each BVC~$i$ responds to boundaries at a preferred distance $d_i$ and direction $\phi_i$ by combining two Gaussian tuning curves over distance and angle. Let 
\[
   \mathbf{r} = [r_1,\dots,r_{n_{\text{res}}}], \qquad 
   \boldsymbol{\theta} = [\theta_0,\dots,\theta_{n_{\text{res}}}]
\]
denote the distances and bearings of the nearest obstacles detected by the $n_{\text{res}}$ LiDAR beams. The firing rate is
\begin{equation}
    v_i^b 
    \;=\; \frac{1}{N_{\text{BVC}}} \sum_{j=0}^{n_{\mathrm{res}}}
    \Biggl(\,
        \frac{\exp\!\bigl[-\tfrac{(r_j - d_i)^2}{2\,\sigma_r^2}\bigr]}{\sqrt{2\pi}\,\sigma_r}
        \;\times\;
        \frac{\exp\!\bigl[-\tfrac{(\theta_j - \phi_i)^2}{2\,\sigma_\theta^2}\bigr]}{\sqrt{2\pi}\,\sigma_\theta}
    \Biggr),
\end{equation}
where $\sigma_r$ and $\sigma_\theta$ control distance and directional tuning, respectively, and $N_{\text{BVC}}$ normalizes activity across the population. The variables are: $r_j$ (distance), $\theta_j$ (bearing) of the $j$-th beam; $d_i$, $\phi_i$ (preferred distance and direction) of BVC~$i$; and $n_{\text{res}}$ (sensor resolution).

\vspace{0.3em}

Several mechanisms for generating multiscale place fields have been proposed \cite{neher2017multiscale,lyttle2013spatialscale}, but boundary-vector-cell input provides the most direct control over field size \cite{barry2006boundary,Scleidorovich2022}. In this model, scale differences arise solely from the BVC tuning widths \(\sigma_r\) and \(\sigma_\theta\), which determine the spatial smoothness of the BVC responses and thus the resolution of downstream place fields. Larger widths produce broader, coarse-scale fields, whereas smaller widths yield more spatially precise fields.

\vspace{1.0em}

\subsubsection{Place Cell Layer}
The place cell (PC) layer is the primary locus of spatial representation in the model, comprising place cells with locally tuned activity in the form of place fields. Each PC receives weighted excitation from BVCs and is inhibited both by total BVC activity (feedforward inhibition) and by other PCs (recurrent inhibition). Together, these interactions produce stable, evenly-distributed place fields  \cite{alabi2020one,alabi2022dissertation,alabi2023rapid}.

\paragraph{Activity Model}
The membrane potential $s_i^p$ of place cell $i$ evolves according to
\begin{equation}
\label{eqn_membrane_potential}
    \tau_p\,\frac{d s_i^p}{d t}
    = -\,s_i^p
    + \sum_{j=0}^{n_b} W_{ij}^{pb}\,v_j^b
    - \Gamma_{pb}\sum_{j=0}^{n_b} v_j^b
    - \Gamma_{pp}\sum_{j=0}^{n_p} v_j^p,
\end{equation}
where $\tau_p$ is the membrane time constant; $W_{ij}^{pb}$ is the BVC$\!\rightarrow$PC synapse (initialized sparsely to promote diverse receptive fields); $v_j^b$ and $v_j^p$ are BVC and PC firing rates; and $\Gamma_{pb},\Gamma_{pp}$ scale feedforward and recurrent inhibition. The firing rate of PC~$i$ is
\begin{equation}
\label{eqn_pc_firing_rate}
    v_i^p
    = \tanh\!\Bigl(\bigl[\psi\,s_i^p\bigr]_{+}\Bigr),
\end{equation}
with rectification enforcing nonnegative output and gain $\psi$ setting response sharpness.

\paragraph{Self-Organization of Place Fields}
When the agent explores a new environment, localized place fields emerge through competitive learning: strongly driven PCs potentiate their BVC inputs, while inhibition suppresses competing cells, thus ensuring that each place cell acquires a distinct place field and the place fields together cover the environment. Synaptic adaptation follows a variant of Oja’s rule~\cite{oja1982simplified}:
\begin{equation}
\label{eqn_ojas}
    \tau_{w^{pb}} \,\frac{d W_{ij}^{pb}}{d t}
    =
    v_i^p
    \Bigl(
        v_j^b
        -
        \tfrac{1}{\alpha_{pb}}\,v_i^p\,W_{ij}^{pb}
    \Bigr),
\end{equation}
where $\tau_{w^{pb}}$ is the learning rate, $\alpha_{pb}$ is a normalization factor, and each synapse is initialized to~1 with probability $p_{pb}=0.25$. The place representations emerging from this mechanism have low redundancy compared to those observed in the actual hippocampus, where place fields can overlap significantly. The choice made in the model represents a tradeoff between redundancy and efficiency, and allows the model to work with a smaller population of place cells. The model can easily accommodate greater redundancy by using localized rather than global inhibitory projections in a layer with many more neurons.

\paragraph{Learning Place Cell Adjacencies}
To create a topological representation of the environment from the place codes, directional adjacency between PCs is encoded in a 3D tensor $W^{pp} = [ W^{pp}_{kij}]$. There are 8 synapses from each PC $j$ to every other PC $i$, one for each of the 8 basis heading directions indexed by $k$. To capture temporal ordering, the model integrates the activity of the presynaptic PC $j$ and the postsynaptic PC $i$ and the activity of each head-direction cell $k$  over a short time window. Integrated activities evolve as:
\begin{equation}
\label{eqn_adjacency_learning_pj}
    \tau_m \frac{d \Upsilon_j^p}{dt} = -\Upsilon_j^p + v_j^p,
\end{equation}
\begin{equation}
\label{eqn_adjacency_learning_pi}
    \tau_m \frac{d \Upsilon_i^p}{dt} = -\Upsilon_i^p + v_i^p,
\end{equation}
\begin{equation}
\label{eqn_adjacency_learning_hk}
    \tau_m \frac{d \Upsilon_k^h}{dt} = -\Upsilon_k^h + v_k^h,
\end{equation}
where $\tau_m$ is the integration constant. These exponential traces preserve a decaying memory of recent activations.

Adjacency weights are updated according to
\begin{equation}
\label{eqn_weight_update}
    \tau_{w^{pp}} \frac{d W_{kij}^{pp}}{dt}
    =
    \Upsilon_k^h \left( v_i^p \Upsilon_j^p - v_j^p \Upsilon_i^p \right),
\end{equation}
where $\tau_{w^{pp}}$ controls learning speed and $\Upsilon_k^h$ gates updates for each head direction. The sign of the difference term encodes the direction of movement: $W_{kij}^{pp}$ thus increases when the agent moves from $j$ to $i$ heading in direction $k$, and decreases when the order is reversed, implementing a temporally smoothed version of spike-timing-dependent plasticity (STDP) \cite{Bi1998}. As a result, the PC layer learns a directed adjacency graph reflecting the topology and navigational flow of the environment.

\begin{figure}[t]
    \centering
    \includegraphics[width=\linewidth]{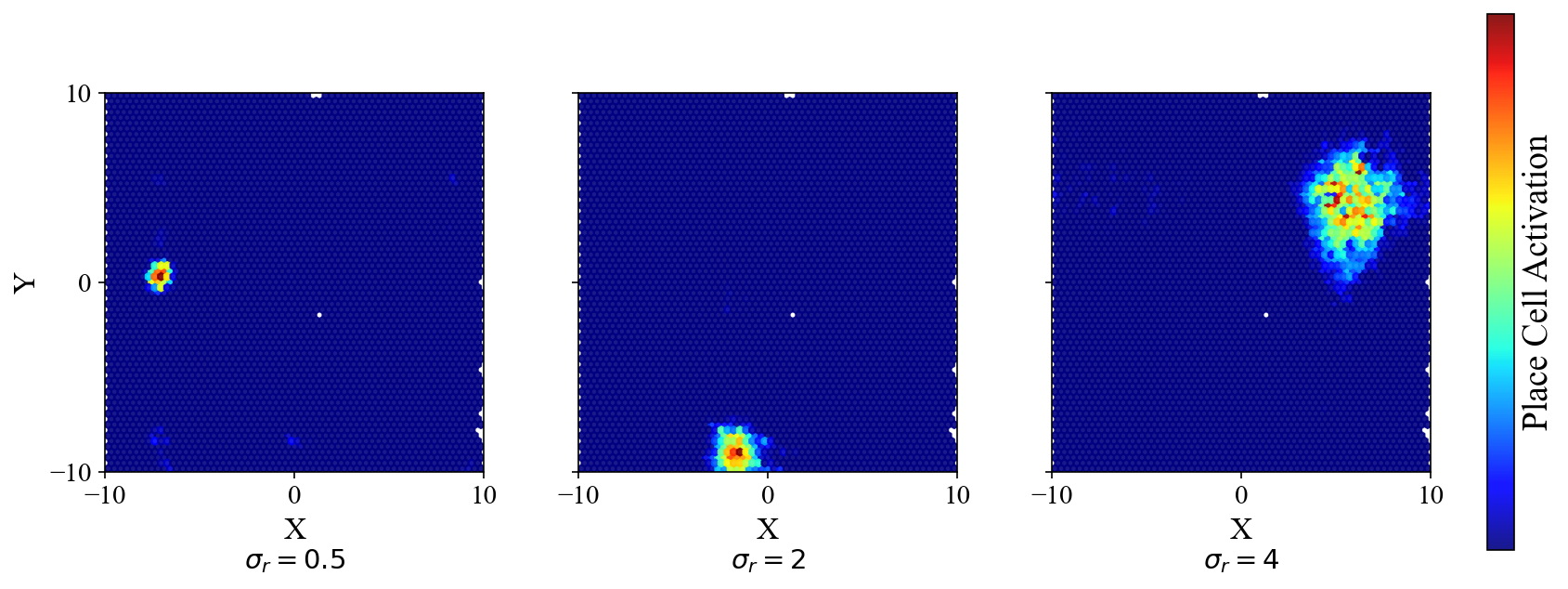}
    \caption{Activation patterns of three representative place cells across spatial scales (\(\sigma_r = 0.5, 2.0,\) and \(4.0\) m) in a \(20 \times 20\)\,m environment. Broader \(\sigma_r\) values produce larger receptive fields, indicating coarser spatial representations. Colors denote activation strength.}
    \label{fig:img_place_cell_scales}
\end{figure}

\paragraph{Multiscale PC Instantiation}
Each spatial scale maintains its own PC population with identical dynamics but distinct upstream BVC tuning widths. Figure~\ref{fig:img_place_cell_scales} shows example receptive fields for the three scales used experimentally.

These scale-dependent receptive fields arise naturally from the BVC tuning parameters; all place-field formation and adjacency learning follow Eqs.~\ref{eqn_pc_firing_rate}–\ref{eqn_weight_update} identically across scales.

\subsubsection{Reward Cell Layer}
\label{reward_cell_layer}
Reward cells become active at the location of rewarded goals \cite{gauthier2018rewardCells,xiao2020rewardPlace}. In the model, each reward cell responds to a single goal. Each scale has a distinct reward cell layer, so that the number of reward cells activated for any goal is equal to the number of spatial scales in the model. Each reward cell for a given scale receives synaptic input from all place cells at that scale. After learning, the reward cell's activity provides a scalar estimate of proximity to the goal as a highly compressed value signal (analogous to ventral striatal or subicular value coding), thus encoding a {\it reward map} peaked at the goal and decreasing monotonically with distance from it in all directions. Because the spatial resolution of each PC scale differs, the resulting reward maps vary in smoothness and spatial extent: coarser scales produce value surfaces with broader support, while finer scales yield more spatially detailed structure.

The reward cell computes a normalized activation based on place cell activity. Let $\mathbf{v}^{p} \in \mathbb{R}^{n_p}$ be the vector of PC firing rates and $\mathbf{w}^{r} \in \mathbb{R}^{n_p}$ the synaptic weights from PCs to the reward cell. The raw activation of the reward cell is:
\begin{equation}
    a^r ({v}^{p}) = \frac{\mathbf{w}^{r} \cdot \mathbf{v}^{p}}{\max\big( \|\mathbf{v}^{p} \|_1, \varepsilon \big)},
\end{equation}
with $\varepsilon = 10^{-4}$ preventing division by zero. The final firing rate is rectified and bounded:
\begin{equation}
    v^r = \min\big( \max( a^r ({v}^{p}), 0), B \big),
\end{equation}
where $B$ is a large upper limit set as a parameter, ensuring nonnegative and biologically plausible activity.

\paragraph{Reward Learning via Replay}
During learning, the reward cell does not use its own forward activation $a^r$ to drive synaptic plasticity. Instead, replay strengthens synapses from place cells that predict or precede reward. When the agent first encounters the goal, PCs active at the goal potentiate their connections onto the reward cell. The model then enters an offline \emph{reverse replay} phase, where previously active PCs are reactivated in time-compressed reverse order, consistent with hippocampal sharp-wave ripple dynamics \cite{foster2006reverse}. Synaptic updates are driven entirely by the replayed place-cell activity, as described below.

This mechanism associates earlier states with eventual reward and supports long-range value assignment. Before replay begins, a weight-update accumulator $\Delta w_i^{r}$ is initialized to zero for all $i$.
During replay, synaptic updates accumulate as
\begin{equation}
\label{eqn_replay}
    \Delta w^{r}_i
    \leftarrow
    \Delta w^{r}_i
    +
    \frac{\overline{v}_i^p}{\|\overline{v}^p\|_{\infty}}\,
    \exp\!\Bigl(-\tfrac{t_r}{\tau_r}\Bigr),
\end{equation}
where $\overline{v}_i^p$ is the replay firing rate of place cell $i$, normalized by $\|\overline{v}^p\|_{\infty}$ to prevent domination by large activations. The variable $t_r$ is a discrete replay-step index ($t_r = 0,1,\dots$), with $t_r=0$ corresponding to the goal state and increasing along the replayed trajectory. The decay time constant $\tau_r$ ensures that states further from the goal contribute progressively less to the accumulated weight update.

After replay concludes, normalized weight updates are applied:
\begin{equation}
\label{eqn_reward_update}
    w^{r}_i
    \leftarrow
    w^{r}_i
    +
    \frac{\Delta w^{r}_i}{\|\Delta \mathbf{w}^{r}\|_\infty},
\end{equation}
enforcing synaptic competition and ensuring that distant or weakly predictive states receive smaller weight changes. The resulting weights encode the reward map: a smooth surface that ideally peaks at the goal and decreases smoothly with distance along the experienced trajectory.

\paragraph{Temporal Difference (TD) Learning}
Replay establishes long-range reward propagation, but the model also refines reward predictions online using a temporal-difference rule. Given an observed reward $R_{\text{next}}$, the prediction error is
\begin{equation}
    \delta = R_{\text{next}} - \mathbf{w}^{r} \cdot \mathbf{v}^{p},
\end{equation}
and synapses update according to
\begin{equation}
    \mathbf{w}^{r} \leftarrow \mathbf{w}^{r} + \eta \delta \mathbf{v}^{p},
\end{equation}
with learning rate $\eta$. TD learning sharpens prediction accuracy near the goal and stabilizes the reward profile across repeated episodes.

\paragraph{Multiscale Reward Maps}
Differences in the upstream BVC tuning widths (\(\sigma_r, \sigma_\theta\)) yield reward maps with complementary spatial properties:
\begin{itemize}
    \item \textbf{Small scale:} High-resolution, rapidly varying reward structure for fine maneuvering near obstacles or the goal, but with limited generalization away from experienced trajectories and limited guidance far from the goal.
    \item \textbf{Medium scale:} Moderately structured reward profiles for intermediate complexity, offering a balance between local detail and spatial generalization.
    \item \textbf{Large scale:} Smooth, slowly varying reward profiles that support long-range guidance, broad spatial generalization, and navigation in open environments.
\end{itemize}

\begin{figure}[t]
    \centering
    \includegraphics[width=\linewidth]{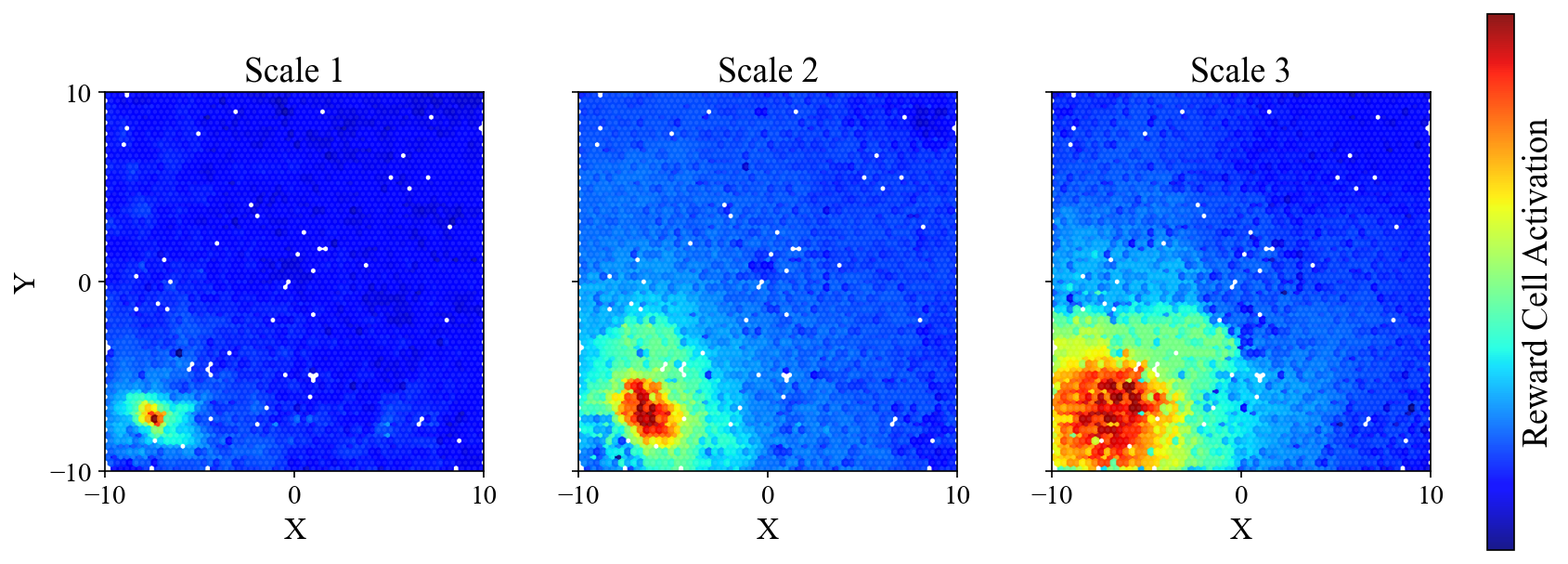}
    \caption{Reward-cell activation patterns for each of the three spatial scales, corresponding to a reward located in the lower-left corner of a \(20 \times 20\)\,m environment. Colors indicate cell activation strength}
    \label{fig:reward_maps}
\end{figure}

These maps (Fig.~\ref{fig:reward_maps}) form the basis for multiscale value fusion during planning: Coarse maps push the agent toward distant goals, while fine maps prevent collisions and refine trajectories locally.

\subsection{Modes of Operation: Exploration and Exploitation}
The model operates in two alternating modes that govern its overall behavior: an \textit{exploration} mode for building spatial representations, and an \textit{exploitation} mode for goal-directed navigation using the learned maps.

\paragraph{Exploration}
During exploration, the agent performs a random walk through the environment to build its internal map using the following processes:
\begin{enumerate}
    \item \textbf{Place Field Formation:}  
    Place cells develop localized receptive fields via competitive learning 
    (Eq.~\ref{eqn_ojas}), gradually tiling the environment.

    \item \textbf{Adjacency Learning:}  
    As the agent moves between place fields, recurrent synapses are updated using directionally gated Hebbian learning 
    (Eqs.~\ref{eqn_adjacency_learning_pj}, \ref{eqn_adjacency_learning_pi}, 
    \ref{eqn_adjacency_learning_hk}, \ref{eqn_weight_update}), forming a tensor of adjacency weights encoding spatial connectivity and movement direction.

    \item \textbf{Reward Map Initialization:}  
    Upon first encountering a goal, the reward cell is activated and synapses from co-active place cells are strengthened, anchoring the reward location in the network, and generating an initial reward map through backward replay (Eq.~\ref{eqn_replay}) and TD learning.
\end{enumerate}

\paragraph{Exploitation}
Once the spatial and reward maps are established, the agent switches to
goal-directed navigation. At each step, it uses internal simulation
(preplay) to imagine the consequences of each possible heading, evaluates
the predicted reward at each scale, fuses these predictions across
scales, and finally selects the optimal continuous movement direction.

\begin{enumerate}
    \item \textbf{Preplay:}
    At each step, the agent performs one-step internal simulations for all basis headings using adjacency-driven predictions:
    \begin{equation}
    \label{eqn_preplay}
        \hat{v}_i^p(\theta)
        = \tanh\!\left(
            \left[
                \sum_{j=1}^{n_p} W^{pp}_{aij} \, v_j^p
                - v_i^p
            \right]_{+}
        \right),
    \end{equation}
    where $W^{pp}_{aij}$ is the slice of the adjacency tensor corresponding to the basis heading $\theta_a$ that is closest to the candidate direction $\theta$. The vector $\hat{\mathbf{v}}^{p}(\theta)$ represents the place-cell activity the agent would expect if it were to move in direction $\theta$. This imagined pattern is used only for evaluating predicted reward across scales; no reward is computed during the preplay step itself. In reinforcement-learning terms, this operation corresponds to a one-step, model-based rollout~\cite{sutton2018reinforcement}.

    \item \textbf{Reward Evaluation and Action Selection via Multiscale Integration:}  
    For each heading, reward predictions from all scales are combined through a dynamic scale-weighting mechanism, and an optimal heading is obtained, as detailed in Sec.~\ref{sec:dynamic_integration}. Scales that fail to produce sufficiently strong reward signals are excluded. It should be noted that the resulting optimal heading need not be one of the 8 basis headings, and actual movement takes place in continuous space.
    
    \item \textbf{Loop Prevention.}
    The agent avoids tight rotational loops by detecting excessive turning without forward motion and temporarily reverting to exploration.
\end{enumerate}

\subsection{Dynamic Integration of Scales}
\label{sec:dynamic_integration}
\begin{figure*}[!t]
    \centering
    \includegraphics[width=0.95\textwidth]{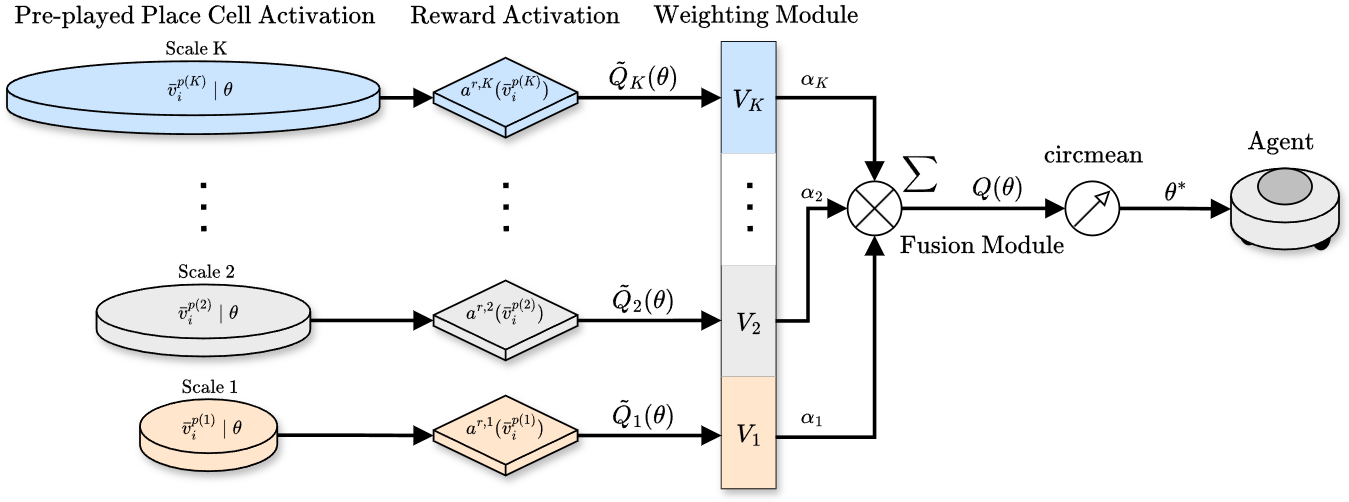}
    \caption{
        Reward fusion architecture. Preplay generates predicted place-cell activity for each heading and scale; directional variation in the reward maps determines the weighting factors \(\alpha_k\), and the weighted sum
        identifies the best action \(a^*\).
    }
    \label{fig:reward_fusion_arch}
\end{figure*}

\paragraph*{Scale-Specific Reward Prediction}
For each scale \(k\), the reward estimate for heading \(\theta\) is obtained by applying the reward-cell activation (Sec.~\ref{reward_cell_layer}) to the preplay-predicted place-cell activity \(\bar{\mathbf{v}}^{p,k}|\theta\) from Eq.~\ref{eqn_preplay}. With reward weights \(\mathbf{w}^{r,k}\),
\begin{equation}
\label{eqn_Qk_compact}
    Q_k(\theta)
    =
    \frac{
        \mathbf{w}^{r,k} \cdot \big(\bar{\mathbf{v}}^{p,k} \mid \theta\big)
    }{
        \max\!\big(
            \|\bar{\mathbf{v}}^{p,k} \mid \theta\|_1,\;
            \varepsilon
        \big)
    }.
\end{equation}

To compare scales based on their directional reward structure rather than magnitude, each profile is normalized:
\begin{equation}
\label{eqn_scale_norm}
    \tilde{Q}_k(\theta)
    =
    \frac{Q_k(\theta)}
         {\max_{\theta'} Q_k(\theta') + \varepsilon}.
\end{equation}

\paragraph*{Variation-Based Scale Weighting}
To prioritize informative scales, the agent measures the total directional reward variation at each valid scale $k$ based on its normalized profile $\tilde{Q}_k(\theta)$:
\begin{equation}
\label{eqn_gradients}
    V_k
    =
    \sum_{d=0}^{n_{\mathrm{hd}}-2}
    \bigl|
        \tilde{Q}_k(\theta_{d+1})
        -
        \tilde{Q}_k(\theta_d)
    \bigr|,
\end{equation}
where $\{\theta_d\}_{d=0}^{n_{\mathrm{hd}}-1}$ are the discrete basis headings used during preplay. These variation values are then normalized across valid scales to obtain the mixing weights used in Eq.~\ref{eqn_combined_reward}:
\begin{equation}
\label{eqn_alpha}
    \alpha_k
    =
    \frac{V_k}{\sum_{j \in \mathcal{V}} V_j + \varepsilon},
\end{equation}
so that scales with larger directional variation $V_k$ receive higher weight in the fused reward $Q(\theta)$.

\paragraph*{Multiscale Fusion}
During exploitation, the normalized reward estimates in each candidate heading are fused across spatial scales:
\begin{equation}
\label{eqn_combined_reward}
    Q(\theta)
    =
    \sum_{k \in \mathcal{V}}
        \alpha_k \,\tilde{Q}_k(\theta),
\end{equation}
where \(\mathcal{V}\) contains only scales whose maximum predicted reward exceeds a validity threshold. The coefficients $\alpha_k$ (defined in Eq.~\ref{eqn_alpha}) specify the relative contribution of each valid scale to the fused reward profile. Scales that do not meet the threshold are excluded from action selection. However, these scales continue updating their place-cell and reward-cell weights, so $\mathcal{V}$ determines only which scales contribute to the \emph{decision}, not which scales continue to \emph{learn}. If no scale is valid ($\mathcal{V} = \varnothing$), the agent briefly switches to exploration before re-evaluating the reward signals.

\paragraph*{Action Selection with Obstacle Avoidance}
To discourage unsafe movements, candidate headings are masked when too close to obstacles. For each heading $\theta_d$, the minimum distance $d(\theta_d)$ to a boundary is estimated from the rangefinder. If $d(\theta_d) < d_{\text{safe}}$, the corresponding reward value is set to zero at all scales before fusion:
\begin{equation}
\label{eqn_obstacle_mask}
    \tilde{Q}_k(\theta_d) \leftarrow 0
    \quad \text{for all } k \text{ if } d(\theta_d) < d_{\text{safe}}.
\end{equation}

The optimal movement direction is then computed as the circular mean of the fused reward profile:
\begin{equation}
\label{eqn_action_angle}
    \theta^*
    =
    \operatorname{arctan2}\!\left(
        \sum_{\theta} Q(\theta)\sin\theta,\;
        \sum_{\theta} Q(\theta)\cos\theta
    \right),
\end{equation}
which yields a single heading in continuous space even when the maximum of $Q(\theta)$ is broad or multi-modal.


\section{Experimental Setup and Performance Analysis}

We evaluated the multiscale model in two settings: (i) \emph{Experiment~1}, which assesses navigational efficiency with pretraining, and (ii) \emph{Experiment~2}, which examines online policy convergence from naive initialization. Table~\ref{tab:two_experiments} summarizes the objectives and procedures.

\begin{table*}[t]
    \centering
    \small
    \caption{Summary of the Two Experiments}
    \label{tab:two_experiments}
    \setlength{\tabcolsep}{4pt}
    \renewcommand{\arraystretch}{1.0}
    \begin{tabular}{p{3cm} p{6cm} p{6cm}}
        \toprule
        & \textbf{Experiment 1: Path Efficiency} & \textbf{\mbox{Experiment 2: Policy Learning}} \\
        \midrule
        \textbf{Objectives} & 
        Evaluate path efficiency for single-scale vs.\ multiscale strategies  
        & 
        Assess online policy convergence and learning dynamics \\
        \midrule
        \textbf{Environments} & 
        Envs~1--4 
        & 
        Env.~1 (open arena) \\
        \midrule
        \textbf{Metrics} & 
        Number of steps to goal 
        & 
        Number of steps to goal per episode, convergence rate over episodes \\
        \midrule
        \textbf{Procedure} & 
        Mapping without reward followed by one replay to form the reward map; fixed start 
        & 
        No pretraining; agent learns place fields, adjacencies, and reward map across episodes; fixed start\\
        \midrule
        \textbf{Test Runs} & 
        20 trials per strategy per environment  
        & 
        51 episodes \(\times\) 5 runs per strategy \\
        \midrule
        \textbf{\mbox{Key Differences}} & 
        Uses pretrained spatial map for evaluation 
        & 
        Fully online learning from Episode~0 \\
        \bottomrule
    \end{tabular}
\end{table*}

\subsection{Experimental Setup}
Simulations were conducted in Webots R2025a \cite{Webots} using a differential-drive robot equipped with a compass for head-direction updates and a planar rangefinder providing \(720\) beams per \(360^\circ\) sweep, which directly fed the BVC layer.

\subsubsection{Experimental vs. Control Groups}
Across both experiments, the multiscale policy served as the experimental condition, differing from the three single-scale control policies only in its integration of spatial scales. All policies were exposed to identical sensory and reward data, ensuring that any performance differences could be attributed directly to the effects of multiscale integration.

\subsubsection{Environment Design}
Four \(20 \times 20\)\,m arenas (Envs.~1–4) with boundary walls and different internal obstacle layouts were used to vary navigational complexity. Obstacles formed open regions and narrow corridors, and the goal was a fixed \(0.5\)-m radius region detected only when reached.

\subsubsection{Navigation Strategies}
The experiments involved three single-scale navigation policies--small, medium, and large--and a multiscale strategy that integrated all three spatial resolutions (Table~\ref{tab:navigation_strategies}).

\begin{table}[t]  
    \centering  
    \renewcommand{\arraystretch}{1.2}  
    \setlength{\tabcolsep}{4pt}  
    \caption{Navigation strategies}
    \label{tab:navigation_strategies}  
    \begin{tabular}{c l c c c}  
        \toprule  
        \textbf{Scale Index} & \textbf{Strategy} & \(\sigma_r\) (m) & \textbf{\# Place Cells} & \textbf{\# PC layers} \\  
        \midrule  
        1 & Small Scale  & 0.5  & 2000 & 1 \\  
        2 & Medium Scale & 2.0  & 500  & 1 \\  
        3 & Large Scale  & 4.0  & 250  & 1 \\  
        -- & Multiscale  & Variable & Variable & 3 \\  
        \bottomrule  
    \end{tabular}  
\end{table}


\subsection{Experiment 1: Path Efficiency Evaluation}
\subsubsection{Overview}
This experiment evaluated goal-directed navigation using a pretrained spatial model in order to isolate the intrinsic effect of place-field scale. All learning processes were completed before evaluation, and no weights were updated during navigation.

\subsubsection{Training Procedure}
Training consisted of two phases. In the \emph{mapping} phase, place fields and directional adjacencies were learned without reward: BVC$\!\rightarrow$PC synapses \(W_{ij}^{pb}\) adapted during exploration, and directional adjacency weights \(W^{pp}_{kij}\) were learned according to the STDP-based rule described earlier, while reward weights \(w_i^{r}\) remained frozen. In the subsequent \emph{goal-seeking} phase, the agent first encountered the goal, triggering a single reverse-replay event that established the reward map via updates to \(w_i^{r}\). After this replay, all weights were frozen for evaluation. Since all three spatial scales were trained in parallel on identical sensory input, the multiscale strategy differed only in how these scale-specific predictions were combined during navigation.

\subsubsection{Evaluation Procedure}
We tested four strategies (small, medium, large, multiscale) across Environments~1–4, running 20 trials per environment. Performance was measured by step count, which fully determines path length under fixed movement increments.

\subsubsection{Results and Figures}
\begin{figure}[t]
    \includegraphics[width=\linewidth]{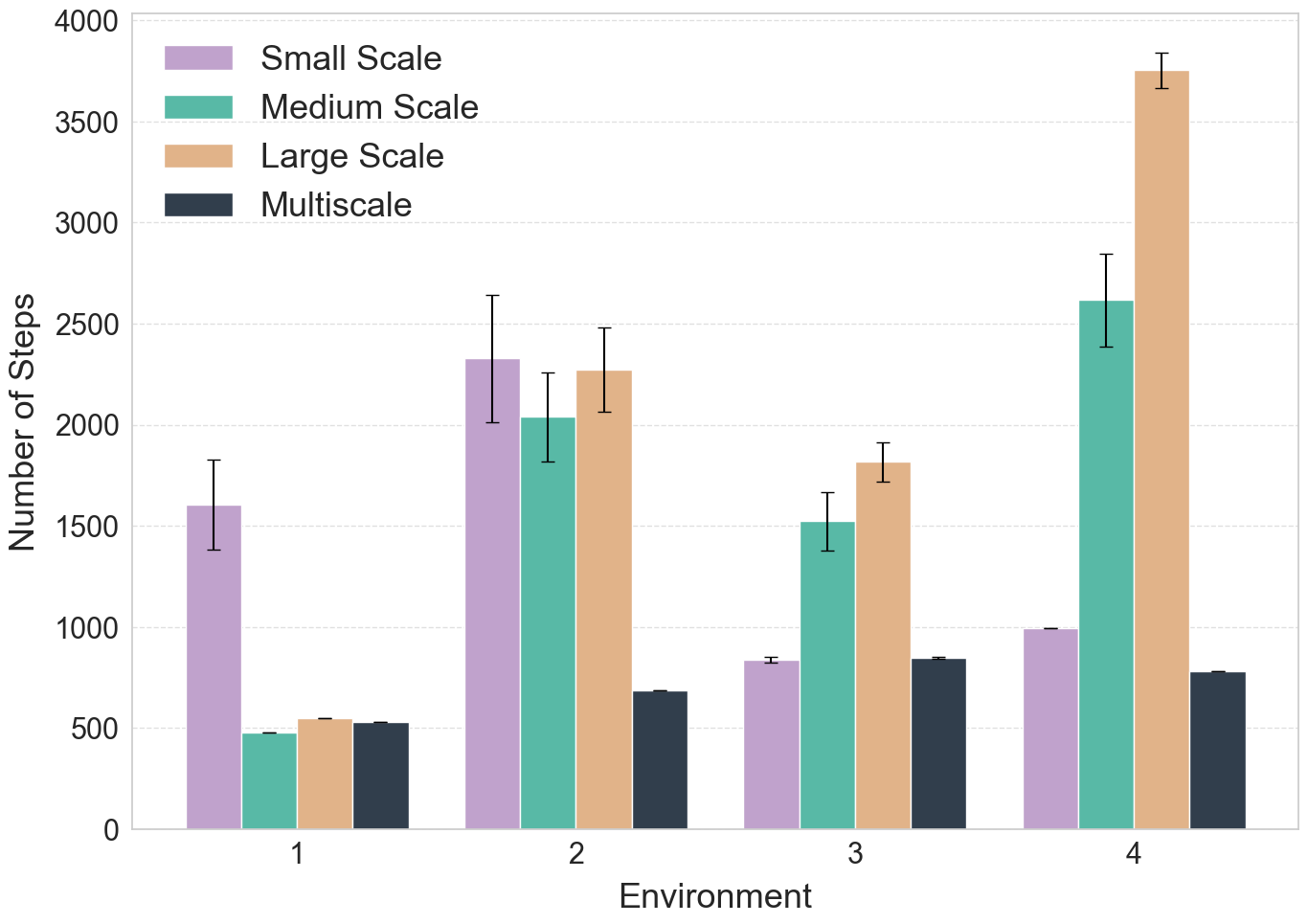}
    \caption{Path efficiency (step count) by strategy and environment. Error bars denote SEM.}
    \label{fig:step_count}
\end{figure}

Figure~\ref{fig:step_count} shows step-count performance for all strategies and environments.

\begin{figure}[t]
    \centering
    \subfloat[Env.~1]{\includegraphics[width=\linewidth]{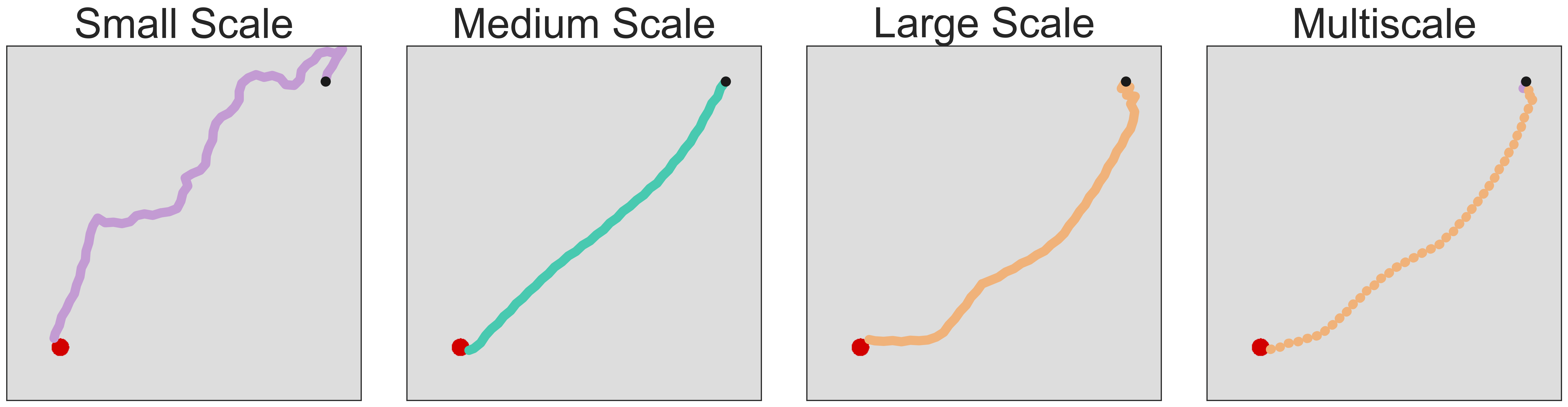}}
    
    \subfloat[Env.~2]{\includegraphics[width=\linewidth]{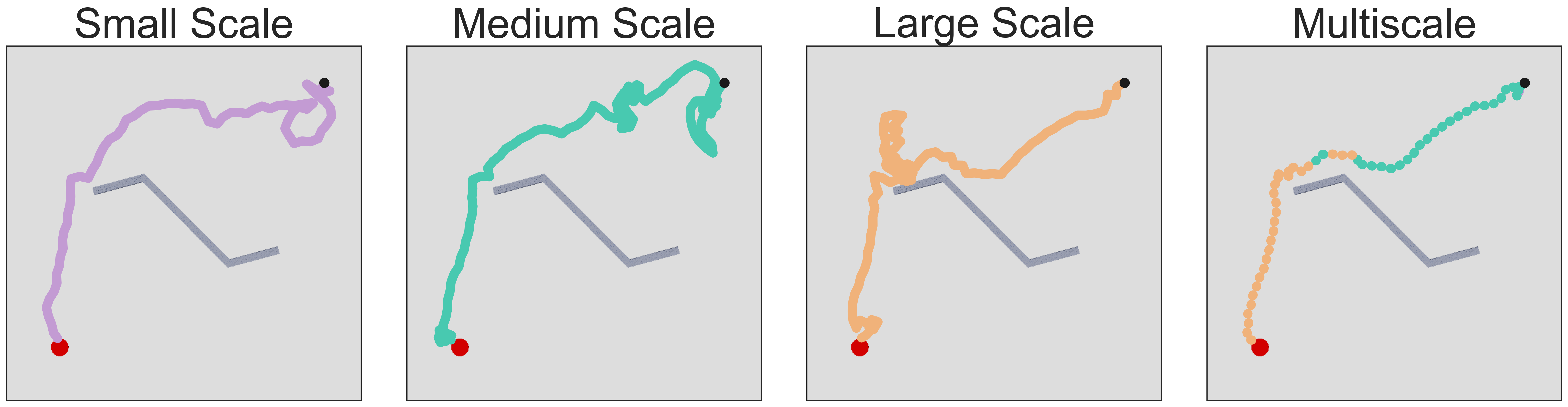}}
    
    \subfloat[Env.~3]{\includegraphics[width=\linewidth]{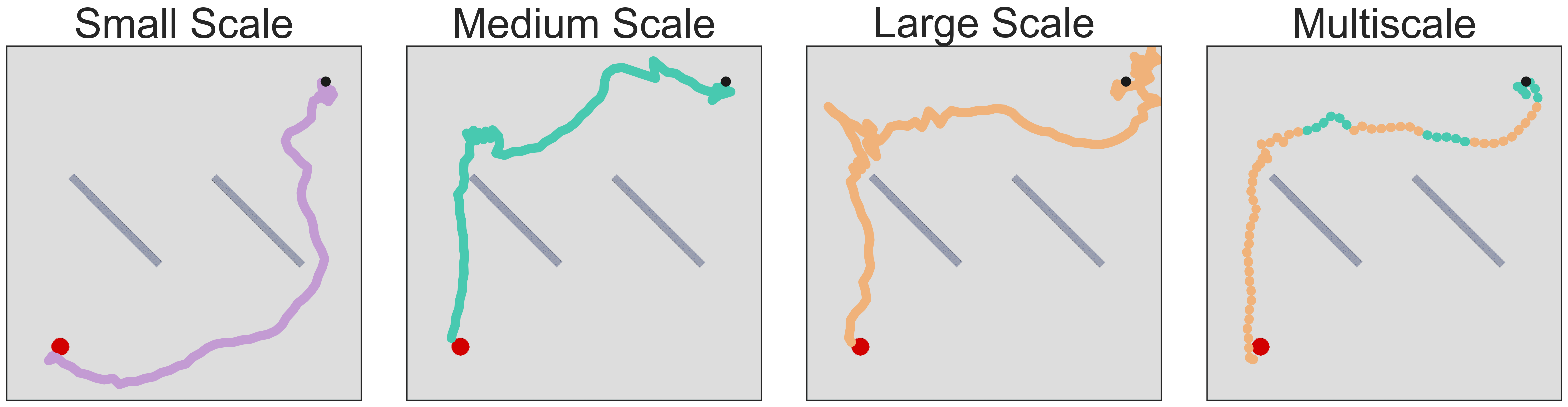}}
    
    \subfloat[Env.~4]{\includegraphics[width=\linewidth]{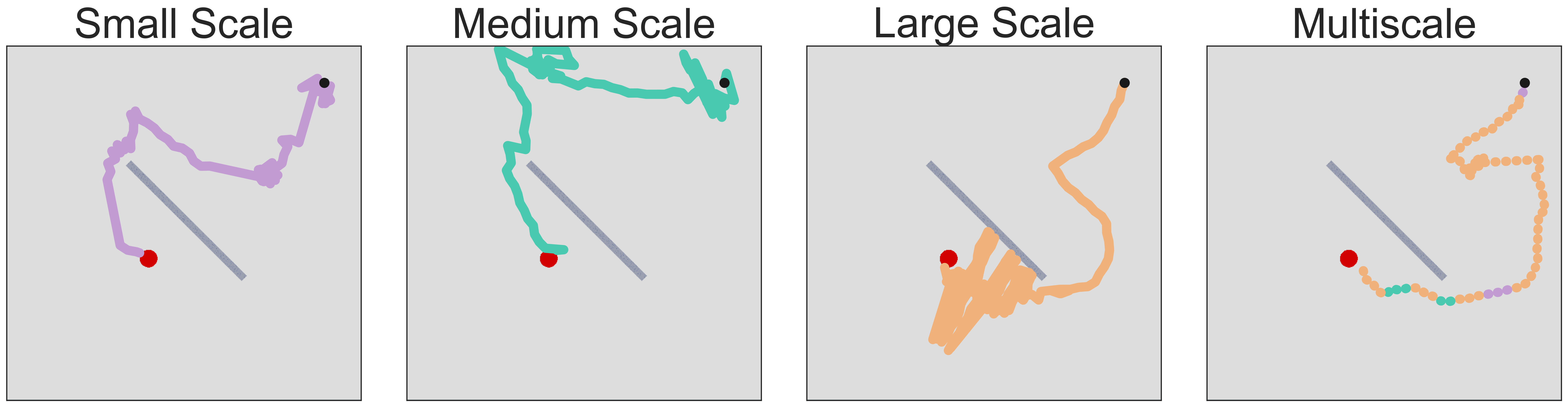}}

    \caption{Randomly-selected trajectories. In multiscale runs, the color along each path segment indicates the scale that dominates decision-making, though scales that do not appear as dominant may still be active.}
    \label{fig:trajectories}
\end{figure}

\subsubsection{Analysis and Discussion}
Figure~\ref{fig:step_count} summarizes step counts across environments.

\textbf{Environment~1 (Open).} 
The medium scale required the fewest steps (480). The small scale performed poorly (1605.35 steps) due to oscillatory corrections. Large and multiscale were similar (547.45 vs.\ 532 steps), with multiscale producing smoother trajectories dominated by coarse-scale activity.

\textbf{Environment~2 (One obstacle).} 
A single barrier magnified scale differences. Small and large scales required similarly long detours (2326.95 and 2273 steps), whereas the multiscale strategy reduced the requirement dramatically (686 steps).

\textbf{Environment~3 (Two obstacles).} 
Small and multiscale showed nearly identical performance (838.90 vs.\ 847.90 steps), though multiscale had lower variance. Medium and large scales required substantially more steps (1522.90 and 1815.55).

\textbf{Environment~4 (Occluded goal).} 
With the goal hidden behind a boundary wall, multiscale again performed best (780 steps). Small scale was next (995 steps), while medium and large scales required much longer detours (2616 and 3752.35 steps).

Across all environments, the multiscale strategy matched or exceeded the strongest single-scale baseline. 
The single-scale results reveal complementary strengths: coarse scales support efficient global guidance in open spaces, while fine scales enable precise local maneuvering in clutter. The multiscale policy leverages these properties by weighting scales according to their directional reward variation at each step, producing consistently shorter and more reliable trajectories with lower across-trial variance.


\subsection{Experiment 2: Policy Learning Evaluation}
\subsubsection{Overview}
Modeled after the Morris water maze task \cite{morris1984developments, vorhees2006morris, Steele1999}, this experiment isolated the learning dynamics in an obstacle-free setting. In the biological paradigm, rodents initially explore the environment without prior knowledge of the platform’s location, and over successive trials develop a stable, goal-directed trajectory. Analogously, the agent in this experiment begins \textit{tabula rasa} and refines its navigation policy through exploration, reward, and replay.

Unlike Experiment~1, which evaluates navigation performance using pretrained spatial representations, Experiment~2 measures how these representations form and stabilize online. Early trajectories are therefore expected to be variable, and the primary focus of this experiment is convergence behavior rather than ideal path efficiency.

\subsubsection{Experimental Procedure}
Each strategy (small, medium, large, and multiscale) was evaluated over 51 episodes (0–50) per trial and 5 trials, yielding 255 episodes per strategy. All evaluations were conducted in Environment~1, an open arena chosen to minimize confounding effects from obstacles and to isolate policy learning. The process consists of two phases:

\begin{itemize}
    \item \textbf{Initial Naive Exploration (Episode~0).}  
    The agent begins with no prior place fields, spatial adjacencies, or reward associations. Navigation is initially random, and all synaptic connections remain plastic, allowing spatial structure to emerge through experience. This phase occurs once at the start of each trial.

    \item \textbf{Reward-Guided Navigation and Policy Refinement (Episodes~1–50).}  
    After the first encounter with the goal, a reverse replay event propagates reward information backward along the experienced trajectory, establishing the internal reward map. Subsequent episodes (starting with Episode 1) use this map to bias movement toward high-value regions while continuing to refine spatial representations and navigation policy.
\end{itemize}

To evaluate learning efficiency, the number of steps required to reach the goal at the end of each episode was recorded. Because Episode~0 and the first few rewarded episodes exhibit transient variability, Episodes~0–5 were excluded from analysis. Mean step count was computed over Episodes~6–50 (45 episodes), providing a stable measure of convergence. If the agent failed to reach the goal within 120 minutes in simulation-time, the episode was terminated and the corresponding step count was recorded.

\subsubsection{Results and Discussion}
\begin{figure}[t]
    \centering
    \subfloat[Small Scale\label{fig:policy_learning_small}]{%
        \includegraphics[width=\linewidth]{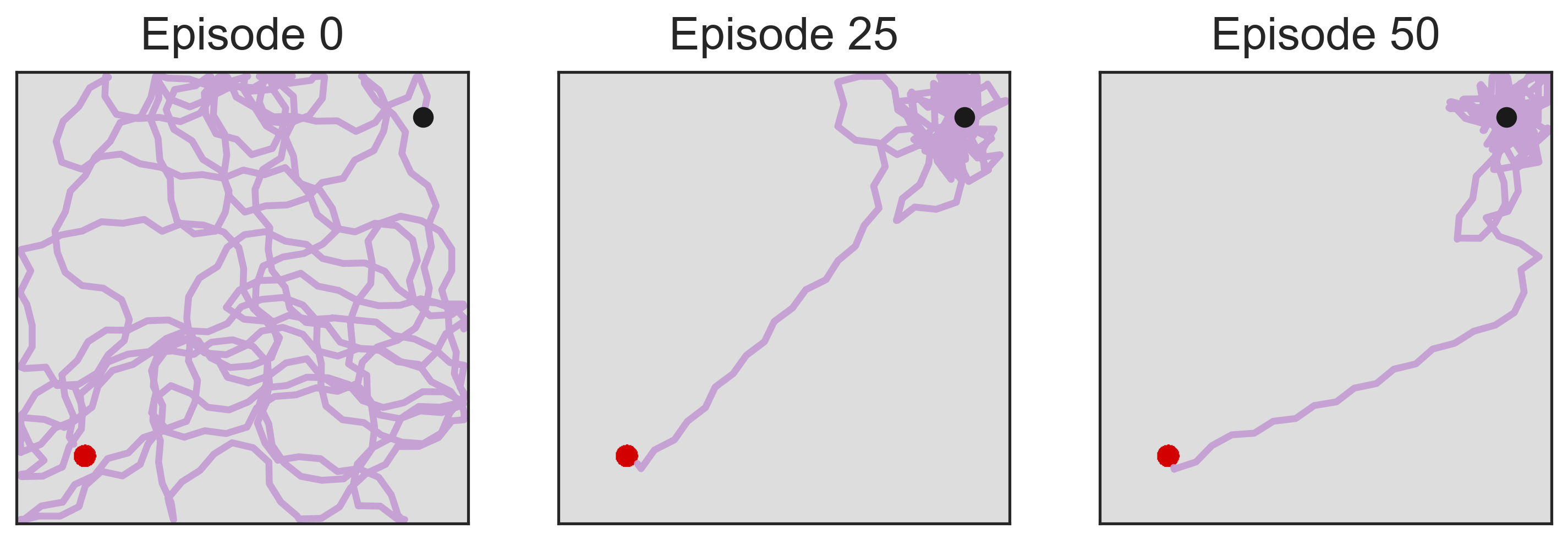}}
    
    \vspace{1.0em}
    
    \subfloat[Medium Scale\label{fig:policy_learning_medium}]{%
        \includegraphics[width=\linewidth]{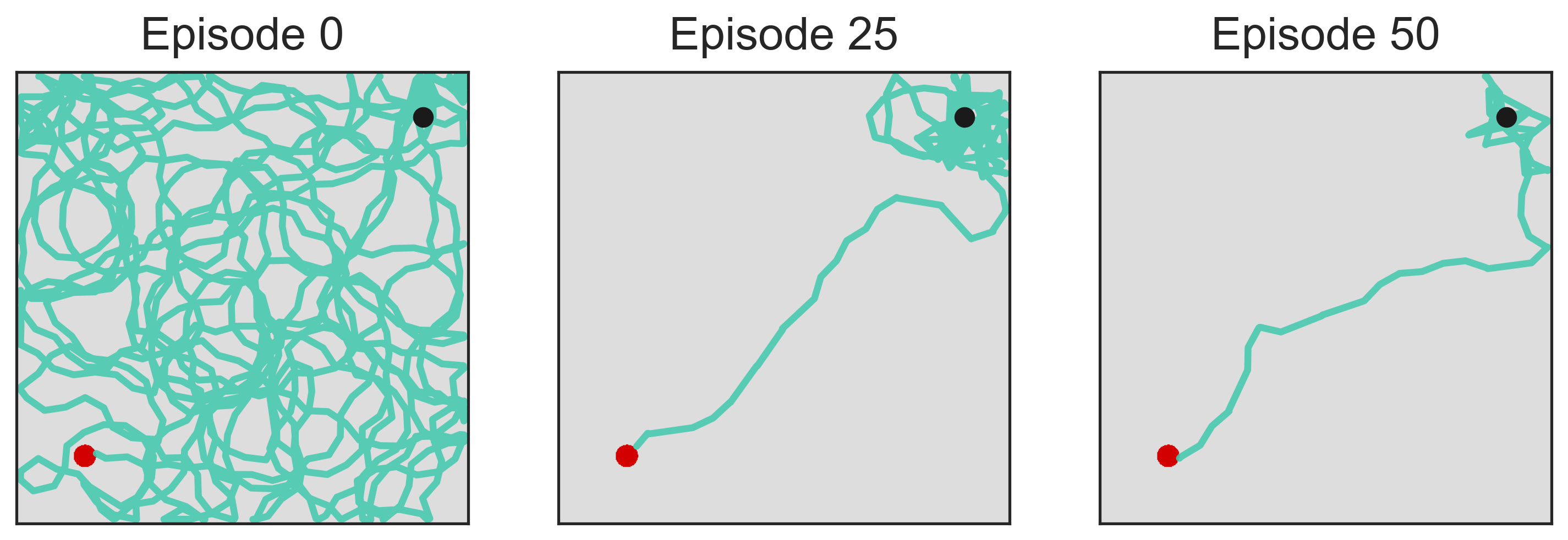}}
    
    \vspace{1.0em}
    
    \subfloat[Large Scale\label{fig:policy_learning_large}]{%
        \includegraphics[width=\linewidth]{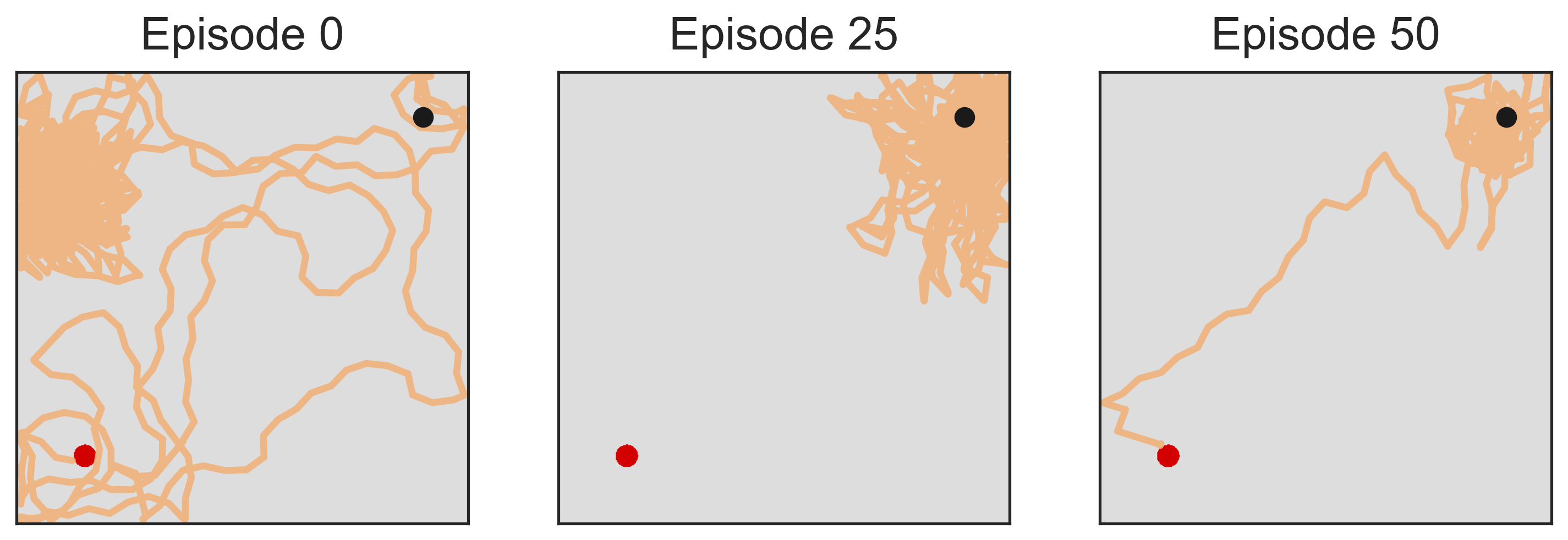}}
    
    \vspace{1.0em}
    
    \subfloat[Multiscale\label{fig:policy_learning_multiscale}]{%
        \includegraphics[width=\linewidth]{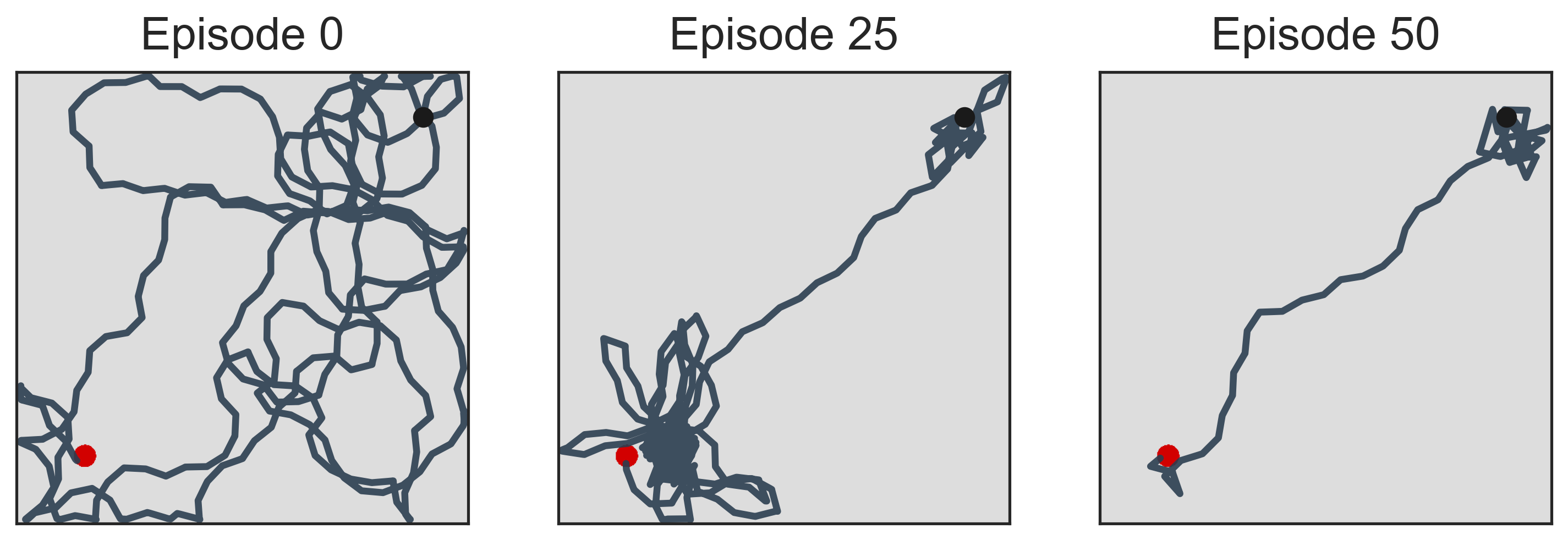}}

    \caption{Paths showing policy refinement over episodes (left$\rightarrow$right).}
    \label{fig:policy_learning}
\end{figure}

Figure~\ref{fig:policy_learning} shows how trajectories evolve across episodes for each strategy. Episode~0 reflects naive goal-seeking exploration before any reward map exists. After the first encounter with the goal, the reward cell's replay mechanism rapidly propagates value information, and trajectories become increasingly direct. Small-scale navigation remains oscillatory (especially when the agent is far from the goal), medium and large scales refine more gradually, and the multiscale agent converges most quickly to smooth, efficient paths.

\begin{figure}[t]
    \centering
    \subfloat[Step-count convergence\label{fig:policy_convergence_a}]{%
        \includegraphics[width=\linewidth]{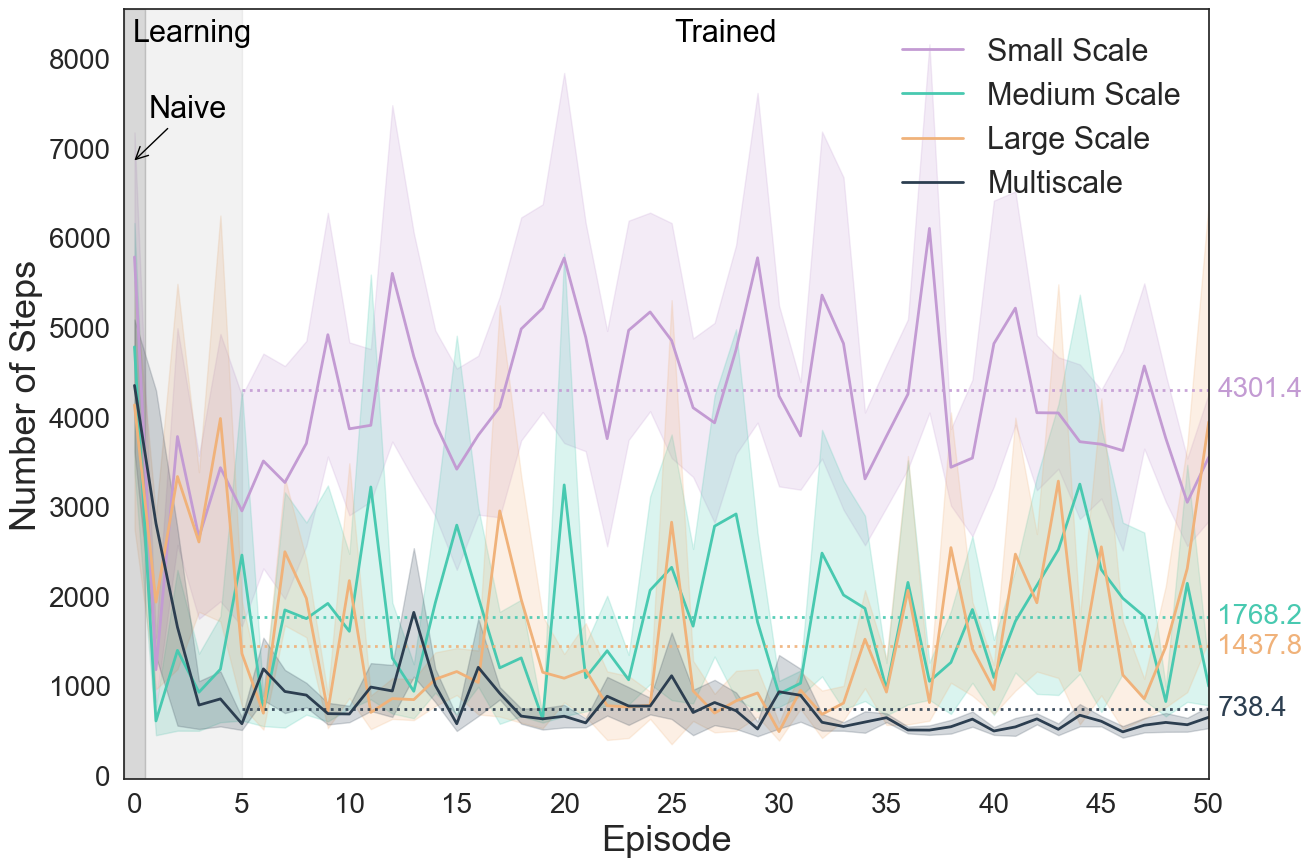}}
    
    \vspace{0.75em}
    
    \subfloat[Rate of change in step count\label{fig:policy_convergence_b}]{%
        \includegraphics[width=\linewidth]{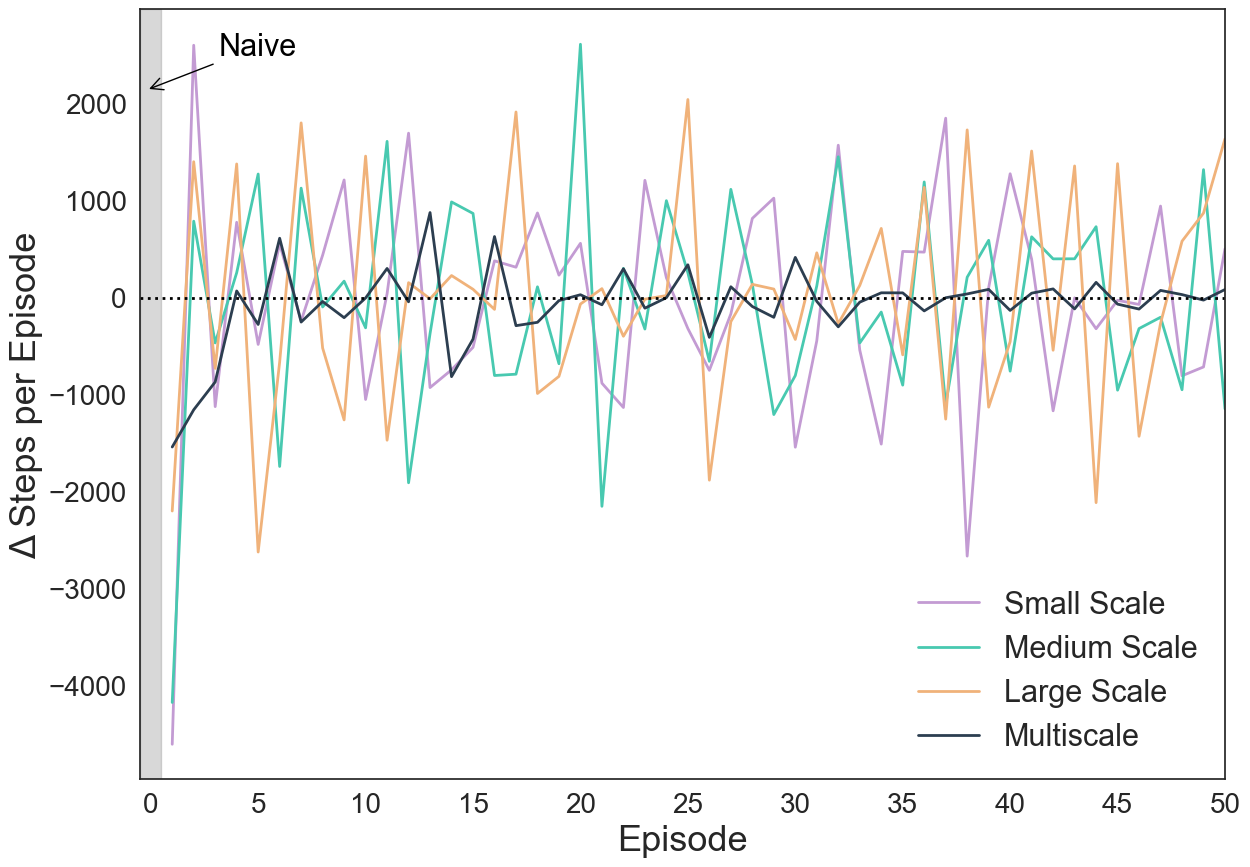}}
    
    \caption{Convergence dynamics across navigation strategies. (a) Step-count convergence with SEM shading; dotted lines show the mean over the final 45 episodes. The Naive, Learning, and Trained episodes are marked by vertical shading. (b) Rate of change in step count ($\Delta$Steps) between successive episodes; the dotted line marks $\Delta$Steps $= 0$ (no change).}
    \label{fig:policy_convergence_combined}
\end{figure}

Figures~\ref{fig:policy_convergence_a} and \ref{fig:policy_convergence_b} quantify these effects. Figure~\ref{fig:policy_convergence_a} shows the number of steps taken by the agent to reach the goal per episode for each navigation strategy, aggregated across trials. Figure~\ref{fig:policy_convergence_b} depicts the rate of change in step count ($\Delta$Steps) reported in Figure~\ref{fig:policy_convergence_a}, computed as the difference between successive episode-wise mean step counts across trials. Values below the zero baseline indicate improvements (fewer steps than the previous episode on average), while values above it indicate regressions. Values below the horizontal baseline ($\Delta$Steps $< 0$) indicate episodes where the agent improves (fewer steps than in the previous episode), whereas values above it indicate regressions.

\textbf{Small Scale.} The small-scale strategy produced the highest final step count ($\sim$4301), reflecting difficulty in acquiring an efficient global navigation policy. Although fine-grained place fields offer precise local cues, they also make the agent highly sensitive to small variations in perceived state. This sensitivity is evident in the $\Delta$Steps trace, which exhibits alternating large negative and positive swings; such volatility indicates frequent over-corrections that repeatedly undo prior improvements.

\textbf{Medium Scale.} The medium-scale strategy achieved a substantially lower final step count ($\sim$1788). Its $\Delta$Steps profile shows moderate fluctuations—primarily negative early in learning, followed by smaller intermittent reversals. This pattern suggests that medium-scale representations support effective global navigation while retaining some inconsistency in fine-grained refinement.

\textbf{Large Scale.} The large-scale strategy converged further ($\sim$1438), consistent with coarse place fields providing strong global guidance in an open environment. The $\Delta$Steps signal contains extended negative dips, corresponding to substantial improvements, interspersed with occasional positive spikes that likely arise from imprecision during final approach to the goal. Compared with smaller scales, the large-scale agent exhibits fewer destabilizing adjustments, though late-episode variability remains present.

\textbf{Multiscale.} The multiscale strategy attained the lowest final step count ($\sim$738) and demonstrated the fastest, smoothest convergence. Its $\Delta$Steps trace remains tightly centered near the zero baseline with noticeably reduced variance after initial learning, reflecting rapid early gains followed by stable policy consolidation. Most $\Delta$Steps values lie close to or slightly below zero.

\paragraph{Statistical and Quantitative Analysis}
A one-way ANOVA confirmed a significant effect of strategy on step count (\(F=127.36,\; p=1.22\times10^{-68}\)). Tukey HSD post-hoc comparisons (Table~\ref{tab:anova_results}) showed that the multiscale strategy significantly outperformed all single-scale strategies (\(p<0.01\)), with the largest difference between multiscale and small scale (3579.4 steps; Cohen’s \(d=0.94\), a large effect). The small-scale condition exhibited the highest SEM (412) compared to multiscale (37), reflecting instability and local overfitting analogous to high variance in myopic RL policies. The non-significant difference between large and medium scales (\(p=0.322\)) suggests overlapping coverage in open environments, where coarse representations dominate global guidance. Collectively, these results demonstrate that dynamic integration of scales enhances convergence efficiency and stability by mitigating fine-scale oscillations and coarse-scale imprecision.

\begin{table}[t]
    \centering
    \caption{ANOVA and Tukey HSD results for step counts.}
    \label{tab:anova_results}
    \resizebox{\linewidth}{!}{
    \begin{tabular}{lccc}
        \toprule
        \textbf{Strategy Comparison} & \textbf{Mean Difference} & \textbf{p-value} & \textbf{Significance} \\
        \midrule
        Large Scale vs. Medium Scale  &  330.4   & 0.322  & No  \\
        Large Scale vs. Multiscale     & -699.4   & 0.0018 & \textbf{Yes}  \\
        Large Scale vs. Small Scale    & 2879.9   & 0.000  & \textbf{Yes}  \\
        Medium Scale vs. Multiscale    & -1029.8  & 0.000  & \textbf{Yes}  \\
        Medium Scale vs. Small Scale   & 2549.5   & 0.000  & \textbf{Yes}  \\
        Multiscale vs. Small Scale     & 3579.4   & 0.000  & \textbf{Yes}  \\
        \bottomrule
    \end{tabular}
    }
\end{table}


\subsection{Summary of Empirical Findings}
Across tasks, the multiscale strategy matched or surpassed all single-scale baselines and converged faster with lower variance. Its adaptive weighting improved robustness in obstacle-rich settings, as supported by ANOVA (\(p=1.22\times10^{-68}\)) and Tukey tests (\(p<0.01\)). These findings show that dynamic scale integration effectively unifies long-range planning with precise local adjustment for robust navigation.


\section{Discussion}

\subsection{Bias--Variance and Reward-Profile Structure}
The results show that each spatial scale contributes distinct statistical properties to the value landscape. Coarser scales generate smooth, low-variance reward profiles that support reliable long-range movement but provide limited detail near obstacles. Finer scales produce sharper directional gradients that enable precise local adjustments but are more susceptible to noise and oscillatory behavior. The variation-based weighting mechanism balances these effects by elevating whichever scale exhibits the clearest directional structure at a given step, without requiring any fixed division of labor across scales. This dynamic selection explains why the multiscale policy yields more consistent trajectories and lower across-trial variance than any single-scale alternative.

\subsection{Parameter Sensitivity and Scale Configurations}
Model performance depends on a small set of interpretable hyperparameters, most notably the BVC tuning widths \(\sigma_r, \sigma_\theta\), which determine place-field size and therefore the spatial frequency content of the reward maps. Extremely small \(\sigma_r\) amplifies noise and can induce zig-zagging, while excessively large \(\sigma_r\) over-smooths gradients and obscures narrow passages. Fusion and gating settings such as the reward-validity threshold and the obstacle mask in Eq.~\eqref{eqn_obstacle_mask} control which scales participate in action selection and how strongly unsafe directions are suppressed. Preplay resolution further trades off computational cost with angular precision. 




\section{Conclusion and Future Work}
This paper presented a navigation model that operates parallel place-field populations at multiple spatial scales and integrates their value estimates using variation-based weighting. In a pretrained path-efficiency evaluation across four arenas, the multiscale policy matched or exceeded the best single-scale baseline in every environment, often yielding shorter, more reliable trajectories with lower across-trial variance. In a separate learning experiment, the same multiscale scheme converged more rapidly and with lower variability than any single-scale control, as reflected in both step-count trajectories and their episode-wise rate of change. Together, these findings support the view that multiscale spatial representations provide complementary advantages, and that fusing them via variation-based weighting promotes more efficient paths and more stable value updates.

Key directions for future work include real-world evaluation, automatic scale discovery, learned scale ratios for fusion, multi-goal, 3D navigation, and integration of grid cells.

\section*{Acknowledgments}

The authors gratefully acknowledge implementation guidance and suggestions from Adedapo Alabi and useful discussions with Dieter Vanderelst.


\end{document}